\documentclass{article}
\usepackage{ijcai20}

\usepackage{times}
\usepackage{soul}
\usepackage{url}
\usepackage[hidelinks]{hyperref}
\usepackage[utf8]{inputenc}
\usepackage[small]{caption}
\usepackage{graphicx}
\usepackage{amsmath}
\usepackage{amsthm}
\usepackage{booktabs}
\usepackage{algorithm}
\usepackage{algorithmic}
\usepackage{graphicx}
\usepackage{tikz}
\usepackage{amsmath}
\usepackage{amssymb}
\usepackage{bm}
\usepackage{xspace}
\usepackage{rotating}
\usepackage{graphics}
\usepackage{subfig}
\usepackage{booktabs}
\usepackage{cancel}

\newcommand{\techname}{\textit{NLocalSAT}\xspace}

\title{NLocalSAT: Boosting Local Search with Solution Prediction}

\author{
Wenjie Zhang$^{1}$\and
Zeyu Sun$^{1}$\and
Qihao Zhu$^{1}$\and
Ge Li$^{1}$\and\\
Shaowei Cai$^{2,3}$\and
Yingfei Xiong$^{1}$\And
Lu Zhang$^{1}$\thanks{Lu Zhang is the corresponding author. The code is available at \url{https://github.com/myxxxsquared/NLocalSAT}}\\
\affiliations
$^1$Key Laboratory of High Confidence Software Technologies (Peking University), MoE;\\
Software Institute, Peking University, China\\
$^2$State Key Laboratory of Computer Science, Institute of Software, Chinese Academy of Sciences, China\\
$^3$School of Computer Science and Technology, University of Chinese Academy of Sciences, China\\
\emails
\{zhang\_wen\_jie, szy\_, zhuqh, lige, xiongyf, zhanglucs\}@pku.edu.cn, caisw@ios.ac.cn
}

\begin{document}

\maketitle

\begin{abstract}
The Boolean satisfiability problem (SAT) is a famous NP-complete problem in computer science. An effective way for solving a satisfiable SAT problem is the stochastic local search (SLS). However, in this method, the initialization is assigned in a random manner, which impacts the effectiveness of SLS solvers. 
To address this problem, we propose \techname.
\techname combines SLS with a solution prediction model, which boosts SLS by changing initialization assignments with a neural network. We evaluated \techname on five SLS solvers (CCAnr, Sparrow, CPSparrow, YalSAT, and probSAT) with instances in the random track of SAT Competition 2018. The experimental results show that solvers with \techname achieve 27\% $\sim$ 62\% improvement over the original SLS solvers.
\end{abstract}

\section{Introduction}

Boolean satisfiability (also referred to as propositional satisfiability and abbreviated as SAT) is the problem to determine whether there exists a set of assignments for a given Boolean formula to make the formula evaluate to true. SAT is widely used in solving combinatorial problems, which are generated from various applications, such as program analysis~\cite{DBLP:conf/popl/HarrisSIG10}, program verification~\cite{DBLP:conf/lpar/Leino10}, and scheduling~\cite{DBLP:conf/icse/KasiS13}. These applications first reduce the target problem into a SAT formula and then find a solution using a SAT solver. However, the SAT problem has proven to be NP-complete~\cite{DBLP:conf/stoc/Cook71}, which means that algorithms for solving SAT Instances may need exponential time in the worst case. Therefore, many techniques have been proposed to increase the efficiency of the search process of SAT solvers.

The state-of-the-art SAT solvers can be divided into two categories, CDCL (Conflict Driven Clause Learning) solvers and SLS (Stochastic Local Search) solvers. CDCL solvers are based on the deep backtracking search, which assigns one variable each time and backtracks when a conflict occurs. On the other hand, SLS solvers initialize an assignment for all variables and then find a solution by constantly flipping the assignment of variables to optimize some score.

Over the last few years, artificial neural networks have been widely used in many problems~\cite{DBLP:journals/corr/EdwardsX16,DBLP:conf/sat/SelsamB19}. A neural network is a machine learning model with a large number of parameters. Neural networks have been used on many data structures, such as sequences~\cite{DBLP:conf/interspeech/MikolovKBCK10}, images~\cite{DBLP:journals/corr/SimonyanZ14a}, and graphs~\cite{DBLP:journals/corr/EdwardsX16}. The graph convolutional network (GCN)~\cite{DBLP:journals/corr/EdwardsX16} is a neural network model on graph structures, which extracts both structural information and information on nodes in a graph. GCN performs well on many tasks on graphs.

There have been some studies on solving SAT problem with neural networks. Some of them use end-to-end neural networks to solve SAT problem directly as the outputs of the neural networks, while others use neural network predictions to boost CDCL solvers.
Selsam et al. proposed an end-to-end neural network model to predict whether a SAT instance is satisfiable~\cite{DBLP:conf/iclr/SelsamLBLMD19} in 2019. Later, Selsam et al. modified NeuroSAT to NeuroCore~\cite{DBLP:conf/sat/SelsamB19}. NeuroCore guides CDCL solvers with unsat-core predictions which are computed every certain interval in the neural network on GPUs. CDCL solvers with NeuroCore solve 6\%-11\% more instances than the original. 

In this paper, we propose \techname, which is the first method that uses a neural network to boost SLS solvers and the first off-line method to boost SAT solvers with neural networks. Different from NeuroCore which induces large overhead to CDCL by an on-line prediction\footnote{The on-line prediction means to predict every certain interval in CDCL.}, \techname uses the prediction in an off-line way. In this method, the neural network is computed only once for each SAT instance.

In our proposed method, we first train a neural network to predict the solution space of a SAT instance. Then, we combine SLS solvers with the neural network by modifying the solvers' initialization assignments under the guidance of the output of the neural network. 
Such combination induces limited overhead, and it can easily be applied to SLS solvers. 
Furthermore, we evaluated SLS solvers, and \techname solves 27\% $\sim$ 62\% more instances than the original SLS solvers. Such experimental results show the effectiveness of \techname.

\paragraph{Contributions.} (1) We train a neural network to predict the solution of a SAT instance. (2) We propose a method to boost SLS solvers by modifying its initialization of assignments with the guidance of predictions of the neural network.
(3) To the best of our knowledge, we are the first to combine SLS with a neural network model and we are the first to propose an off-line method to boost SAT solvers with a neural network.

\section{Approach}

\begin{figure}
    \centering 
    \includegraphics[width=\linewidth]{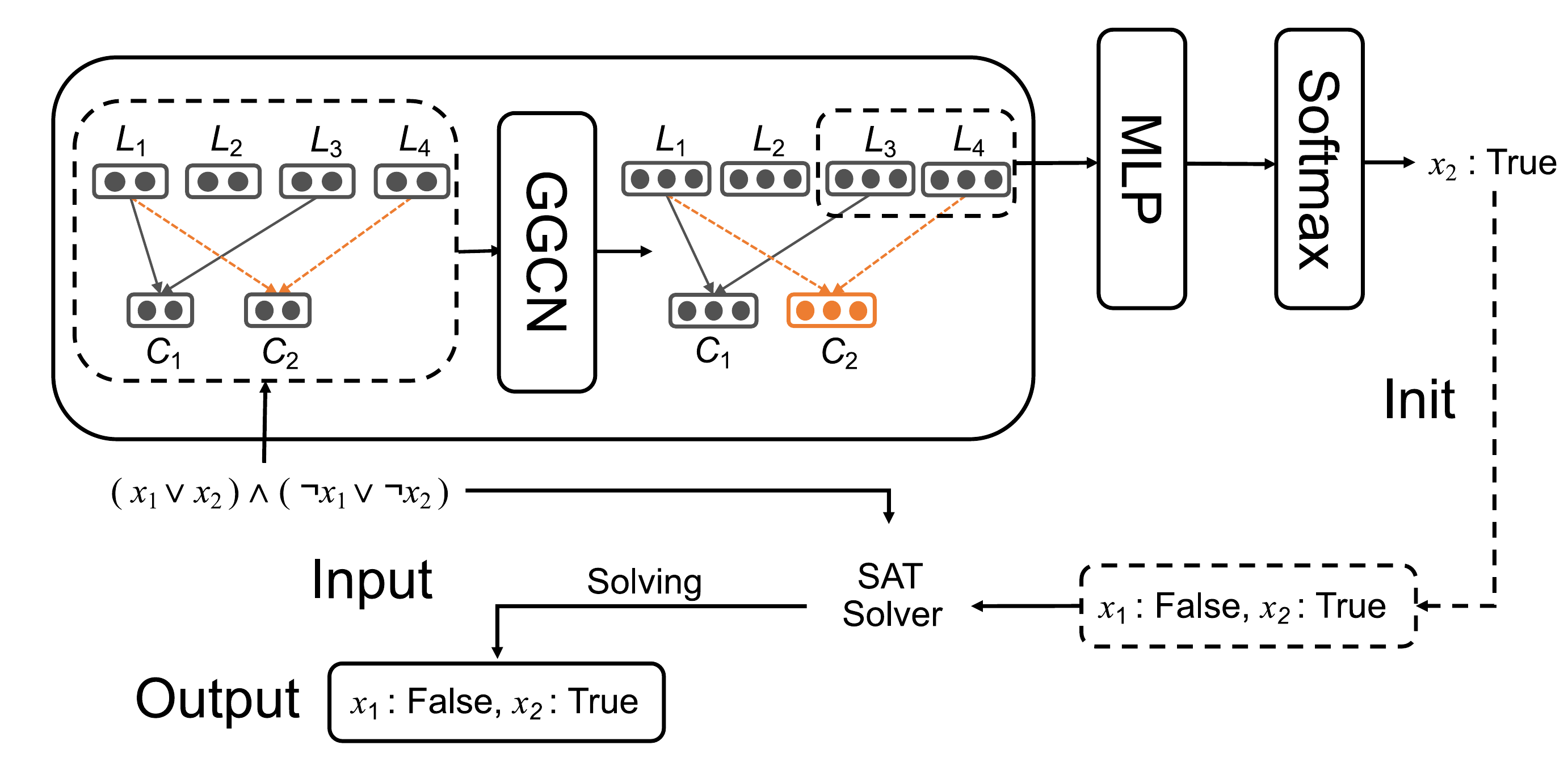}
    \caption{The overview of our model, \techname.}
    \label{fig:overview}
\end{figure}

Figure~\ref{fig:overview} shows an overview of our model, \techname. Our model combines a neural network and an SLS solver. Given a satisfiable input formula, the neural network is to output a candidate solution and the solver is to find a final solution with the guidance of the neural network. For an input formula, we first transfer it into a formula graph, which is further fed to a graph-based neural network for extracting features. Then the neural network in \techname outputs a candidate solution for the formula by a multi-layer perceptron after the neural network. We use this candidate solution to initialize the assignments of an SLS SAT solver to guide the search process.

\subsection{Formula Graph}
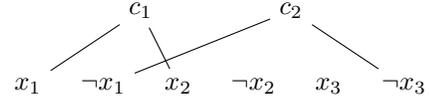
\begin{figure}[t]
    \centering
    \begin{tikzpicture}
    \node (c1) at (-1,1) {$c_1$};
    \node (c2) at (1,1) {$c_2$};
    \node (x1) at (-2.5,0) {$x_1$};
    \node (nx1) at (-1.5,0) {$\neg x_1$};
    \node (x2) at (-0.5,0) {$x_2$};
    \node (nx2) at (0.5,0) {$\neg x_2$};
    \node (x3) at (1.5,0) {$x_3$};
    \node (nx3) at (2.5,0) {$\neg x_3$};
    \draw (c1) edge (x1)
    (c1) edge (x2)
    (c2) edge (nx1)
    (c2) edge (nx3);
    \end{tikzpicture}
    \caption{The bipartite graph representation for the CNF formula.}
    \label{fig:bipartite}
\end{figure}

To take the structural information of an input formula into consideration, we first transfer it into a formula graph.
A general Boolean formula can be any expressions consisting of variables, conjunctions, disjunctions, negations, and constants. All Boolean formulas can be reduced into an equisatisfiable conjunctive normal form (CNF) with linear length in linear time~\cite{tseitin1968complexity}. In a CNF, a SAT instance is a conjunction of clauses $C_1 \wedge C_2 \wedge \cdots \wedge C_n.$ Each clause is a disjunction of literals (i.e., variables and negated variables) $C_i = L_{i1} \vee L_{i2} \vee \cdots \vee L_{in},$ where $L_{ij} = x_k$ or $L_{ij} = \neg x_k$. In this paper, we assume that all SAT problems are in CNFs.
A SAT instance $S$ in the CNF can be seen as a bipartite graph $G = (C, L, E)$, where $C$ (clause set of $S$) and $L$ (literal set of $S$) are the node sets of $G$ and $E$ is the edge set of $G$. $(c, l)$ is in $E$ if and only if the literal $l$ is in the clause $c$.
$A$ is the adjacent matrix of the bipartile graph $G$. The element $A_{ij}$ of the adjacent matrix equals to one when there is an edge between node $i$ and node $j$, otherwise 0. For example, $\left(x_1 \lor x_2\right) \land \neg \left(x_1 \land x_3\right)$ is a Boolean formula. This Boolean formula can be converted into $\left(x_1 \lor x_2\right) \land \left(\neg x_1 \lor \neg x_3\right)$ in the conjunctive normal form. The bipartite graph for this problem is shown in Figure~\ref{fig:bipartite}. The adjacency matrix for this graph is
$$
A = \left(\begin{matrix}
1 & 0 & 1 & 0 & 0 & 0\\0 & 1 & 0 & 0 & 0 & 1
\end{matrix}\right).
$$

\subsection{Graph-Based Neural Network}

The graph-based neural network aims to predict the candidate solution for a SAT instance. The network consists of a gated graph convolutional network to extract structural information about the graph and a two-layer perceptron to predict the solution.

\subsubsection{Gated Graph Convolutional Network} Inspired by NeuroSAT~\cite{DBLP:conf/iclr/SelsamLBLMD19}, we use a similar gated graph convolutional network (GGCN) to extract features of variables. The gated graph convolutional network (GGCN) takes the adjacency matrix as the input and outputs the features of each variable extracted from the graph.

In a SAT instance, the satisfiability is not influenced by the names of clauses and literals (e.g., the satisfiability of two formulas $\left(x_1 \lor x_2\right)$, $\left(x_3 \lor x_4\right)$). To use this property, for an input formula graph $G$, we initialize each clause $c_i \in G$ as a vector $\bm{c}^\text{(init)} \in \mathbb{R}^d$, each literal $l_i \in G$ as another vector $\bm{l}^\text{(init)} \in \mathbb{R}^d$, where $d$ is the embedding size and $d$ is set to 64 in this paper. These vectors are further fed to GGCN to extract structural information in the graph.

Each iteration of GGCN is an update for the vectors of these nodes where each node updates its vector by taking its neighbors' information (vectors). Formally, at the $t$-th iteration, the detailed computations for clause $c$ and literal $l$ are represented by
\begin{equation}
    \bm c_{t} = \mathrm{LSTMCell}(\bm c_{t-1}, \sum_{l' \in G} \tilde{A}_{cl'} \bm l_{t-1}')
\end{equation}
\begin{equation}
    \bm l_{t} = \mathrm{LSTMCell}(\bm l_{t-1}, \sum_{c' \in G} \tilde{A}_{c'l} \bm c_t' + \neg \bm l_{t-1})
\end{equation}
Here, $\tilde{A}$ is a normalized adjacency matrix for the graph (the detailed computation is presented in Equation~\ref{eq:norm}). $\bm l'_t$ and $\bm c'_t$ denote the vector of the literal $l'$ and clause $c'$ at the $t$-th iteration. $\neg \bm l_{t}$ is the vector of negated literal of $l$ at the $t$-th iteration. $\bm c_0$ = $\bm c'_0$ = $\bm{c}^\text{(init)}$, $\bm l_0$ = $\bm l_0'$ = $\bm{l}^\text{(init)}$. The $\mathrm{LSTMCell}$ is a long short-term memory (LSTM) unit with layer normalization. We use symmetrical normalization on adjacency matrix.
\begin{equation}
    \tilde{A} = S_1^{-1/2} A S_2^{-1/2},
    \label{eq:norm}
\end{equation}
where $S_1$ and $S_2$ are the diagonal matrices with summation of $A$ in columns and rows.

We apply the GGCN layer of 16 iterations on the initial value and get a vector containing structural information about each literal. Then, the two vectors for a literal and its negation are concatenated for each variable.

\subsubsection{Two-layer Perceptron} After GGCN, vectors of nodes contain structural information of literals. A two-layer perceptron  $\mathrm{MLP}$ with hidden 512 size is applied on the vector for each variable to extract classification information from the structural information. Through a softmax function, we get the probability for the variable to be true.
\begin{equation}
\begin{aligned}
&P(v=\mathrm{FALSE}), P(v=\mathrm{TRUE})\\
= &\mathrm{softmax}\left\{\mathbf{W}_2~\mathrm{ReLU}\left[\mathbf{W}_1(\mathrm{\mathbf l_{v}: \mathbf l_{\neg v}})+\mathbf{b}_1\right]+\mathbf{b}_2\right\},
\end{aligned}
\end{equation}
where $\mathbf W_1, \mathbf W_2, \mathbf b_1, \mathbf b_2$ are weights and biases for the two perceptrons and the colon indicates the connection of two vectors.

\subsubsection{Loss Function} Our model is trained by minimizing the cross-entropy loss against the ground truth. For the predicted variables $<v_1,v_2\cdots,v_n>$, where $n$ denotes the number of variables, the cross-entropy loss is computed as
\begin{equation}
    \begin{aligned}Loss =  - \sum_{i=1}^{n}&[g(v_i)\log(P(v=\mathrm{TRUE}))\\
    + &\left(1-g(v_i)\right)\log(P(v=\mathrm{FALSE}))],\end{aligned}
\end{equation}
where $g(v_i)$ denotes the ground truth of $v_i$.

\subsection{Training Data}

The goal of our network is to predict the solution to the SAT instances, so we should generate training data for SAT instances with solution labeling. Due to the scarcity of SAT Competition data, using additionally generated small SAT instances could help provide thorough training. We generate two training datasets and train our model in order. Our model is first pretrained with a large number of generated small SAT instances. We generate tens of millions of small SAT instances with solution labeling. Such large amounts of data can make the training process more effective and avoid overfitting. These small instances can help our network better learn structural information. Then, our model is fine-tuned on a dataset generated from SAT Competitions. By finely tuning on the dataset from SAT Competition instances, our model can learn specific information in the field and learn to predict the solution on large instances.

\subsubsection{Small Generated Instances}

We generate small instances containing 10 to 20 variables by a random generator. The number of clauses is 2 to 6 times the number of variables. Each clause contains 3 variables with the probability of 0.8 and 2 variables with the probability of 0.2. Each clause is generated by randomly randomly selecting variables or negated variables from the instance.

After the generation of instances, we solve these instances with a SAT solver. We drop unsatisfied instances because our model only learns to predict the solution of a satisfiable SAT instance. However, there can be more than one solution for a specific SAT instance, which can confuse the neural network. To address this problem, we use a complete solver to find all solutions for each SAT instance. Then we label each variable based on whether itself or its negation appears more among all solutions.

After data generation, we pretrain \techname on these small instances. Small data instances have high overhead during training the neural network, so we combine several small instances into one large batch. Since our model is independent of the order and the size of SAT variables, we can combine many instances via simply putting them together.

\subsubsection{SAT Competition Instances}

To train our model on larger instances, we generate training data from SAT instances in random tracks of SAT Competitions.
We use a SAT solver to get one solution for each instance.
Instances that are not solved by the solver will be removed. We then use the solution as labeling in the training data.

Training on GPU requires more memory than evaluation. So, if the instance is too large to fit in the memory of our GPU during training, we will cut these instances into smaller ones. Let us denote the largest number of variables that can fit into the memory of our GPU as $N_L$. For every large instance, we first get a solution $S_0$ with a SAT solver. Then, we sample $N_L$ variables $X_0$ from all variables. For one clause $c$ in the original SAT instance, if $c$ contains no literals from $X_0$, the clause is removed. If $c$ is not satisfied on $X_0$ after removing literals that are not from $X_0$, the clause is also removed. Clauses with only one literal are also removed to prevent the instance from being too easy. Otherwise, the clause $c$ remains in the instance. If the sampling generates a instance with too few clauses, the instance is removed because this will lead to too many solutions.

\subsection{Combination with Local Search}

Stochastic local search algorithms can be considered as an optimization process. The solver flips the assignments of variables to maximize some total score. For example, we can use the number of clauses that evaluate to true as the score. When the score reaches the number of clauses, the instance is solved. Algorithm \ref{alg:sls} shows a general algorithm in a local search solver.

\begin{algorithm}[tb]
\caption{Algorithm for stochastic local search solvers}\label{alg:sls}
\textbf{Data}: SAT instance $P$
\begin{algorithmic}
    \WHILE{End condition not reached}
        \STATE $S \leftarrow$ initialize assignment randomly;\\
        \WHILE{Restart condition not reached}
            \IF{$P$ evaluate to true under $S$}
                \STATE Return $S$;
            \ENDIF
            \STATE $l \leftarrow$ Select a variable by some heuristics;\\
            \STATE Flip($S$, $l$);
        \ENDWHILE
    \ENDWHILE
\end{algorithmic}
\end{algorithm}

In SLS solvers, it is not counter-intuitive that the initial assignment has a great impact on whether it can quickly find a solution, because there can be many local optimum in the instance and badly initialized assignments near a local optimum can cause the SLS solvers to get stuck. Intuitively, the closer to the solution of the instance, the easier it is to find a solution. In order to avoid falling into the local optimum, these solvers restart the searching process by reassigning new random values to variables after a period of time without a score increase. However, most of the existing SLS solvers initialize assignments in a random manner. Random generation of initial values can explore more space for the SAT instance. However, if the distance between initial values and solution values is too large, the solver is more likely to fall into a local optimum.

We propose a new initialization method using the output of our neural network. Before starting the solver, we run our neural network to predict a solution. We replace the initialization function with our neural initialization, where, instead of randomly generating 0 or 1, the function assigns the predicted values to a variable with a probability of $p_0$ and assigns the negation of the predicted value with a probability of $1-p_0$. This neural-initialization function can keep the assignment with a probability of $p_0$ to explore near the candidate solution and explores new solution space with a probability of $1-p_0$ in case that the neural network's prediction is wrong. The initialization process is shown in Algorithm \ref{algo:init}. Algorithm \ref{alg:nlocalsat} shows the architecture of our modified SLS solvers. Note that our neural network model is executed only once for one SAT instance. Though the cost of computing a neural network is high, the cost of calling a neural network only once is acceptable, which consumes 0.1 seconds to tens of seconds depending on the size of the instance.

\begin{algorithm}[tb]
\caption{Initialization of variables with \techname}\label{algo:init}
\textbf{Data}: Probability $p_0$, Assignment of variables $\mathrm{assignment}$, Neural network predictions $N$
\begin{algorithmic}
    \FOR{$i < \mathrm{number~of~variables}$}
        \IF{$\mathrm{rand}() < p_0$}
            \STATE $\mathrm{assignment}[i] \leftarrow N[i]$
        \ELSE
            \STATE $\mathrm{assignment}[i] \leftarrow \neg N[i]$
        \ENDIF
    \ENDFOR
\end{algorithmic}
\end{algorithm}

\begin{algorithm}[tb]
\caption{Algorithm for stochastic local search solvers with \techname (The underlined code is the different part from the original solvers)}\label{alg:nlocalsat}
\textbf{Data}: SAT instance $P$
\begin{algorithmic}
    \STATE \underline{$N \leftarrow $ predicting solution of $P$ by \techname;}
    \WHILE{End condition not reached}
        \STATE \underline{$S \leftarrow$ initialize assignment with $N$;}
        \WHILE{Restart condition not reached}
            \IF{$P$ evaluate to true under $S$}
                \STATE Return $S$;
            \ENDIF
            \STATE $l \leftarrow$ Select a variable by some heuristics;
            \STATE Flip($S$, $l$);
        \ENDWHILE
    \ENDWHILE
\end{algorithmic}
\end{algorithm}

\section{Experiments}

\subsection{Datasets}

Our model was trained on a dataset with generated instances with small SAT instances (denoted as $Dataset_{small}$) and a dataset with instances in random tracks of SAT Competitions in 2012, 2013, 2014, 2016, 2017 (denoted as $Dataset_{comp}$) \footnote{The competition of SAT in 2015 was called SAT Race 2015. There was no random track in SAT Race 2015}. Our model was evaluated on instances in the random track of SAT Competition in 2018 (denoted as $Dataset_{eval}$) with 255 SAT instances in total. We found that there are several duplicate instances in 2018 and previous years, so we removed them from the training and validation datasets to ensure instances in the $Dataset_{eval}$ are generated with different random seeds with those in $Dataset_{comp}$. So, it's almost impossible to have isomorphic instances between these two datasets. However, there will be some similar substructures between the training set and the test set, so that neural networks can predict by learning these substructures.

The $Dataset_{comp}$ and the $Dataset_{eval}$ both contain two categories of instances, i.e., uniformly generated random SAT instances (denoted as \textit{Uniform})~\cite{belov2014generating} and hard SAT instances generated with a predefined solution (denoted as \textit{Predefined})~\cite{DBLP:conf/socs/BalyoC18}.

\subsection{Pretraining}

In $Dataset_{small}$, we generated about $2.5 \times 10^7$ small instances and combined them into about $4 \times 10^5$ batches with about ten thousand variables each as our pretraining dataset. We generated 200 batches in the same approach with different random seeds as validation data during pretraining.

We trained our model to converge using the Adam~\cite{DBLP:journals/corr/KingmaB14} optimizer with its default parameters by minimizing the loss function. After pretraining, the precision on the validation dataset of $Dataset_{small}$ is 98\%.

\subsection{Training}

We used $Dataset_{comp}$ as the training dataset and the validation dataset. We loaded the pretrained model and continued to train with the same optimizer and loss function. After training, the precision on the validation dataset of $Dataset_{comp}$ is 95\%.

\subsection{Evaluation}

We tested our proposed method on five recent SLS solvers, i.e., CCAnr~\cite{DBLP:conf/sat/CaiLS15}, Sparrow~\cite{DBLP:conf/sat/BalintF10}, CPSparrow, YalSAT~\cite{biere2016splatz}, probSAT~\cite{balint2018probsat}. These solvers have performed very well among SLS solvers on random tracks of SAT Competitions in recent years. CCAnr is an SLS solver proposed in 2015 to capture structural information on SAT instances. CCAnr is a variant of CCASat~\cite{DBLP:journals/ai/CaiS13}. CCAnr performs better on all tracks of SAT Competitions than CCASat. Sparrow is a clause weighting SLS solver. CPSparrow is a combination of Sparrow and a preprocessor Coprocessor~\cite{DBLP:conf/sat/BalintM13}. CPSparrow is the best pure SLS solver in the random track of SAT Competition 2018. YalSAT is the champion of the random track of SAT Competition 2017.

Due to the strong randomness of SLS solvers, the experiments for SLS solvers were performed three times with three different random seeds and then we aggregated the results.

We evaluated these original solvers and those modified with \techname on $Dataset_{eval}$. We also evaluated three other solvers MapleLCMDistChronoBT~\cite{ryvchin2018maple_lcm_dist_chronobt}, gluHack, and Sparrow2Riss~\cite{balint2018sparrowtoriss} under the same set. MapleLCMDistChronoBT and Sparrow2Riss are the champions of SAT Competition 2018 in the main track and the random track. gluHack is the best CDCL solver of SAT Competition 2018 in the random track. MapleLCMDistChronoBT is a CDCL solver with recently proposed techniques to improve performance such as chronological backtracking~\cite{DBLP:conf/sat/NadelR18}, learned clause minimization~\cite{DBLP:conf/ijcai/LuoLXML17}, and so on. Sparrow2Riss is a combination of Coprocessor, Sparrow, and a CDCL solver Riss.

We set up a timeout limit to 1000 seconds. Solvers failed to find a solution within the time limit will be killed immediately. In our experiments, $p_0$ is set to $0.9$. Our experiments were performed on a work station with an Intel Xeon E5-2620 CPU and a TITAN RTX GPU. During our experiment, the time for initialization of the GPU environment was ignored but the time of the GPU computation was included in the total time.

\section{Results}

\subsection{Number of Instance Solved}

Table \ref{tab:num-problems} shows the number of instances solved in 1000 seconds time limit. Each row represents a tested solver. The experiments of SLS solvers are performed three times to reduce the randomness of results. Each number in the rows of SLS solvers is the average and the standard deviation of results in the three experiments. Each column in the table represents a category of instances in the dataset. The number in parentheses indicates the total number of instances in the category. $Dataset_{eval}$ contains unsatisfiable instances~\cite{belov2014generating}. However, no solvers reported unsatisfiability on any instances within the time limit. So, the solved problems in Table \ref{tab:num-problems} are all satisfiable ones.

\begin{table}
    \centering
    \resizebox{\columnwidth}{!}{
    \begin{tabular}{lcccc}
    	\toprule
    	Solver & Predefined($165$) & Uniform($90$) & Total($255$)\\
    	\midrule
    	CCAnr & $107.3\pm1.2$ & $18.0\pm0.8$ & $125.3\pm1.2$\\
    	CCAnr with \techname & $165.0\pm0.0$ & $12.7\pm0.9$ & $\mathbf{177.7\pm0.9}$\\\midrule
    	Sparrow & $126.7\pm0.5$ & $23.7\pm1.7$ & $150.3\pm1.2$\\
    	Sparrow with \techname & $165.0\pm0.0$ & $31.0\pm0.8$ & $\mathbf{196.0\pm0.8}$\\\midrule
    	CPSparrow & $128.0\pm0.8$ & $27.0\pm1.6$ & $155.0\pm1.4$\\
    	CPSparrow with \techname & $165.0\pm0.0$ & $32.0\pm0.8$ & $\mathbf{197.0\pm0.8}$\\\midrule
    	YalSAT & $75.0\pm0.0$ & $49.5\pm1.5$ & $124.5\pm1.5$\\
    	YalSAT with 	\techname & $165.0\pm0.0$ & $37.3\pm0.9$ & $\mathbf{202.3\pm0.9}$\\\midrule
    	probSAT & $75.7\pm0.5$ & $51.0\pm0.0$ & $126.7\pm0.5$\\
    	probSAT with 	\techname & $165.0\pm0.0$ & $40.7\pm1.2$ & $\mathbf{205.7\pm1.2}$\\\midrule
    	Sparrow2Riss & $165$ & $23$ & $188$\\
    	gluHack & $165$ & $0$ & $165$\\
    	MapleLCMDistBT & $165$ & $0$ & $165$\\
    	\bottomrule
    \end{tabular}
    }
    \caption{Number of instances solved in time limit.}\label{tab:num-problems}
\end{table}

The experimental result shows that solvers with \techname solve more instances than the original ones. CCAnr, Sparrow, CPSparrow, YalSAT, and probSAT with \techname solve respectively 41\%, 30\%, and 27\%, 62\%, and 62\% more instances than the original solvers. This improvement has been shown in \textit{Predefined} instances and in \textit{Uniform} instances on Sparrow and CPSparrow. Sparrow with \techname and CPSparrow with \techname solve more instances than all other solvers including the champions on SAT Competition 2018. 
Sparrow2Riss is a combination of a preprocessor, an SLS solver, and a CDCL solver, thus showing good performance, but the SLS solvers with \techname still outperforms Sparrow2Riss. CDCL solvers perform well on \textit{Predefined} instances and \techname can help to improve performance particularly on this category, from which we can conclude that \techname can improve particularly on those instances, on which CDCL solvers perform well.

\subsection{Time of Solving Instances}

\begin{figure}[htb]
    \centering
    \subfloat[CCAnr]{\includegraphics[width=.5\linewidth]{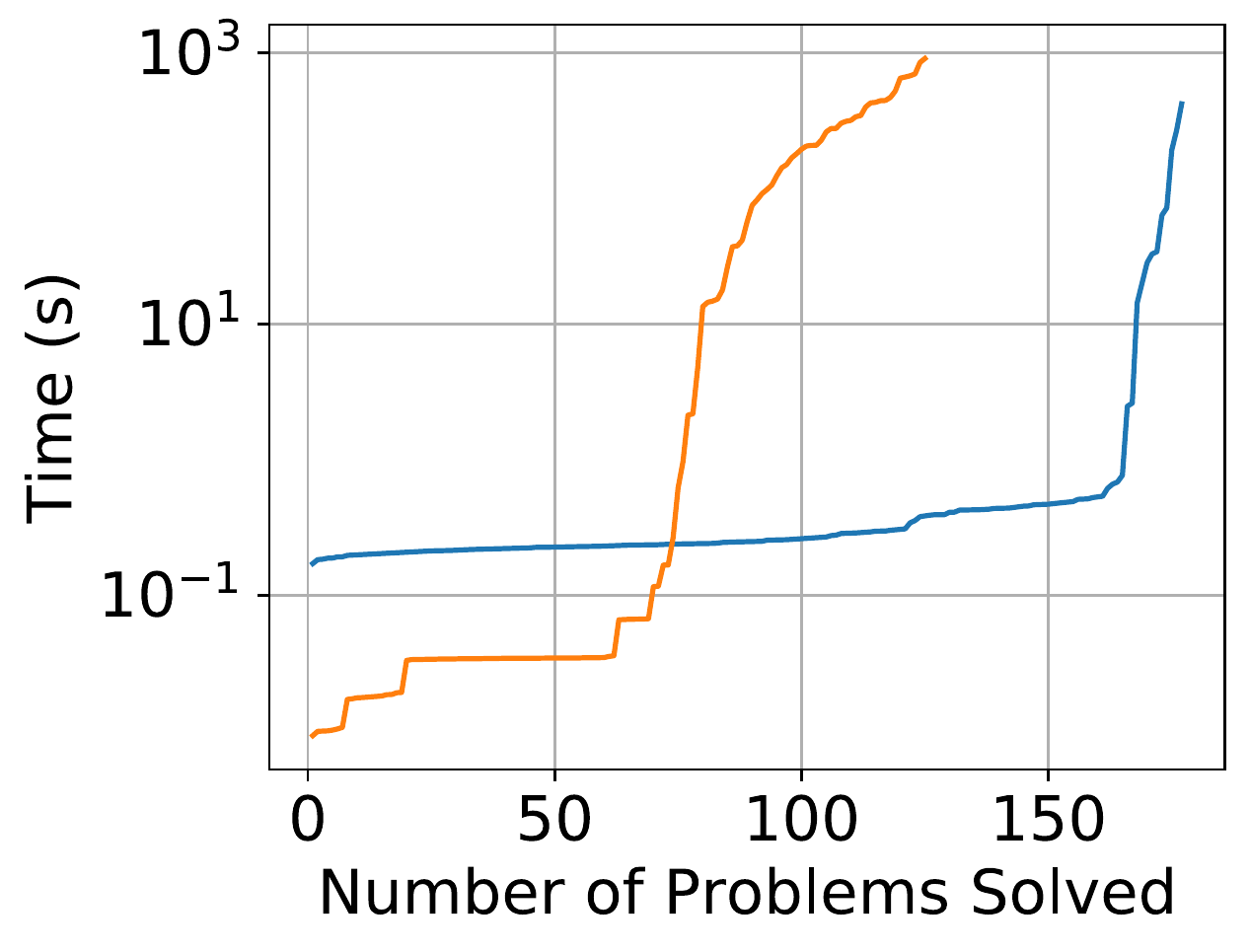}}
    \subfloat[Sparrow]{\includegraphics[width=.5\linewidth]{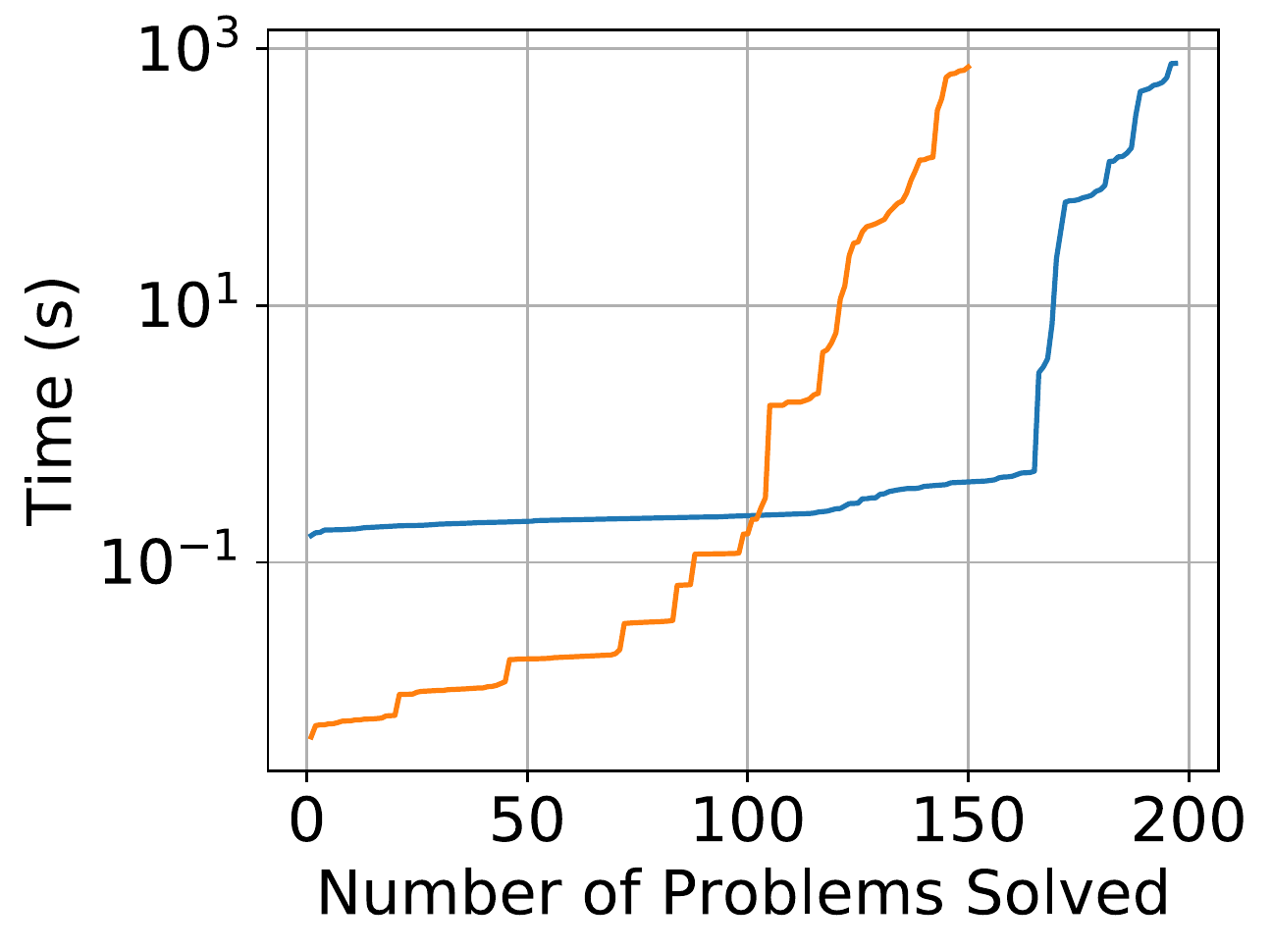}}

    \subfloat[CPSparrow]{\includegraphics[width=.5\linewidth]{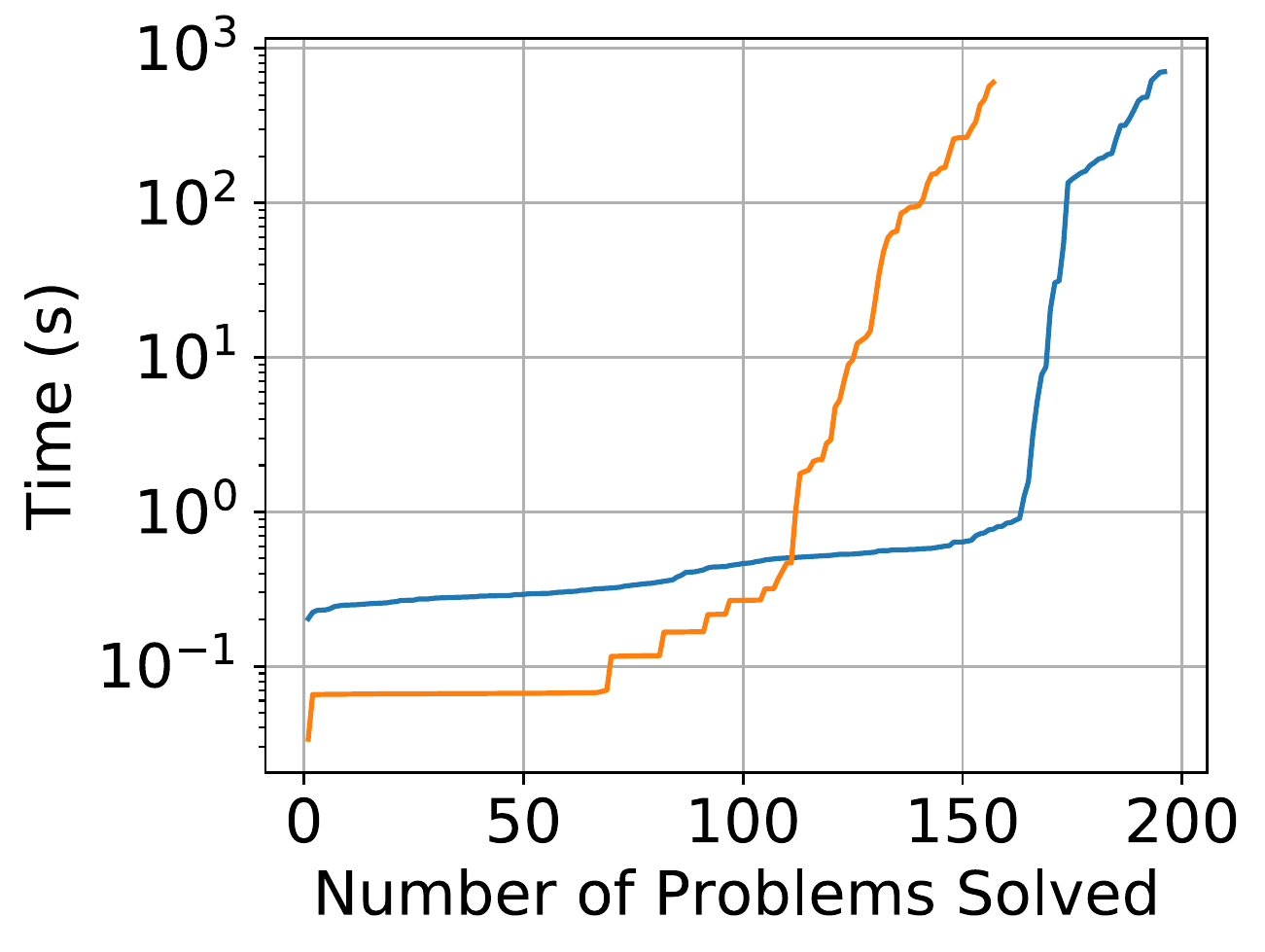}}
    \subfloat[YalSAT]{\includegraphics[width=.5\linewidth]{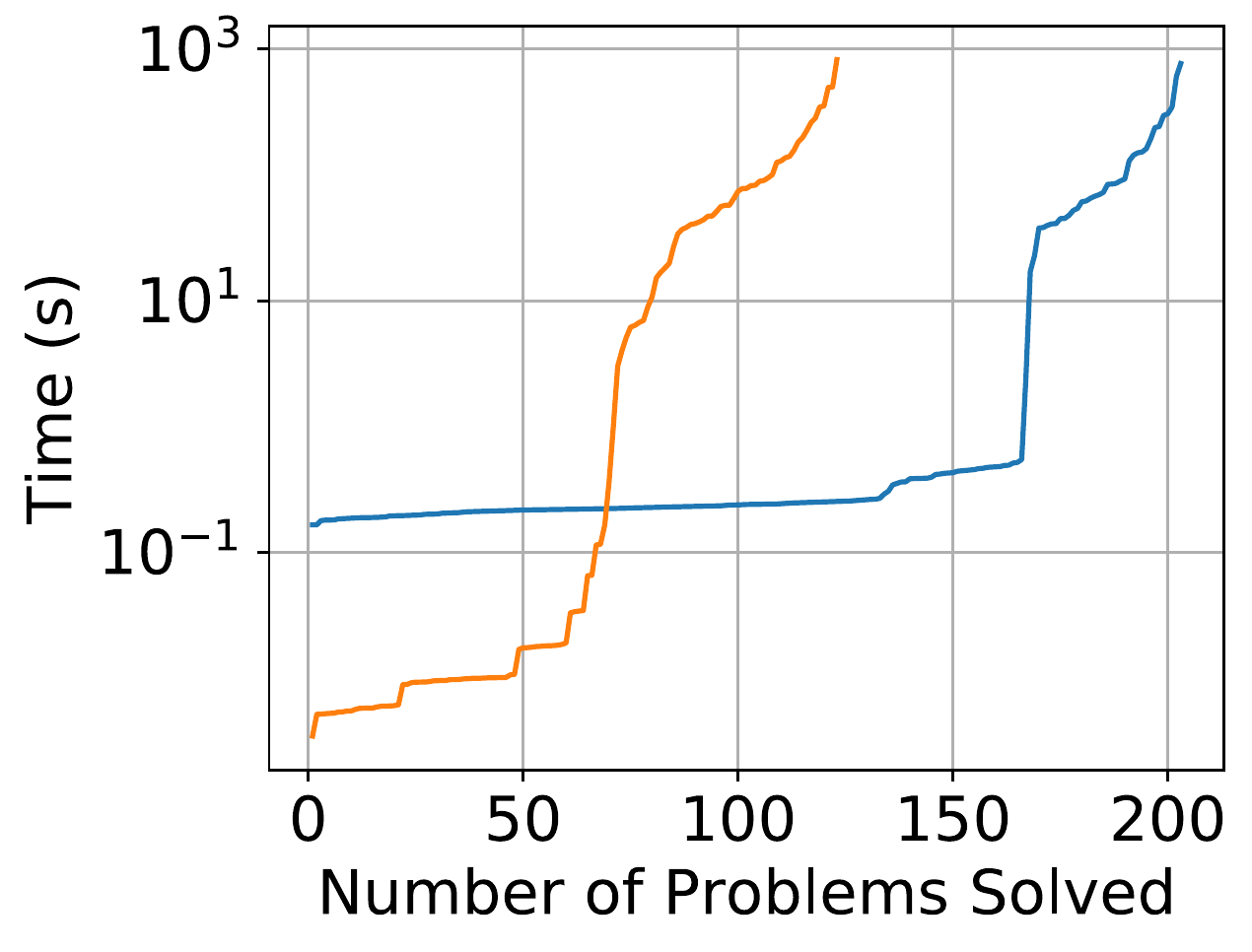}}
    
    \subfloat[probSAT]{\includegraphics[width=.5\linewidth]{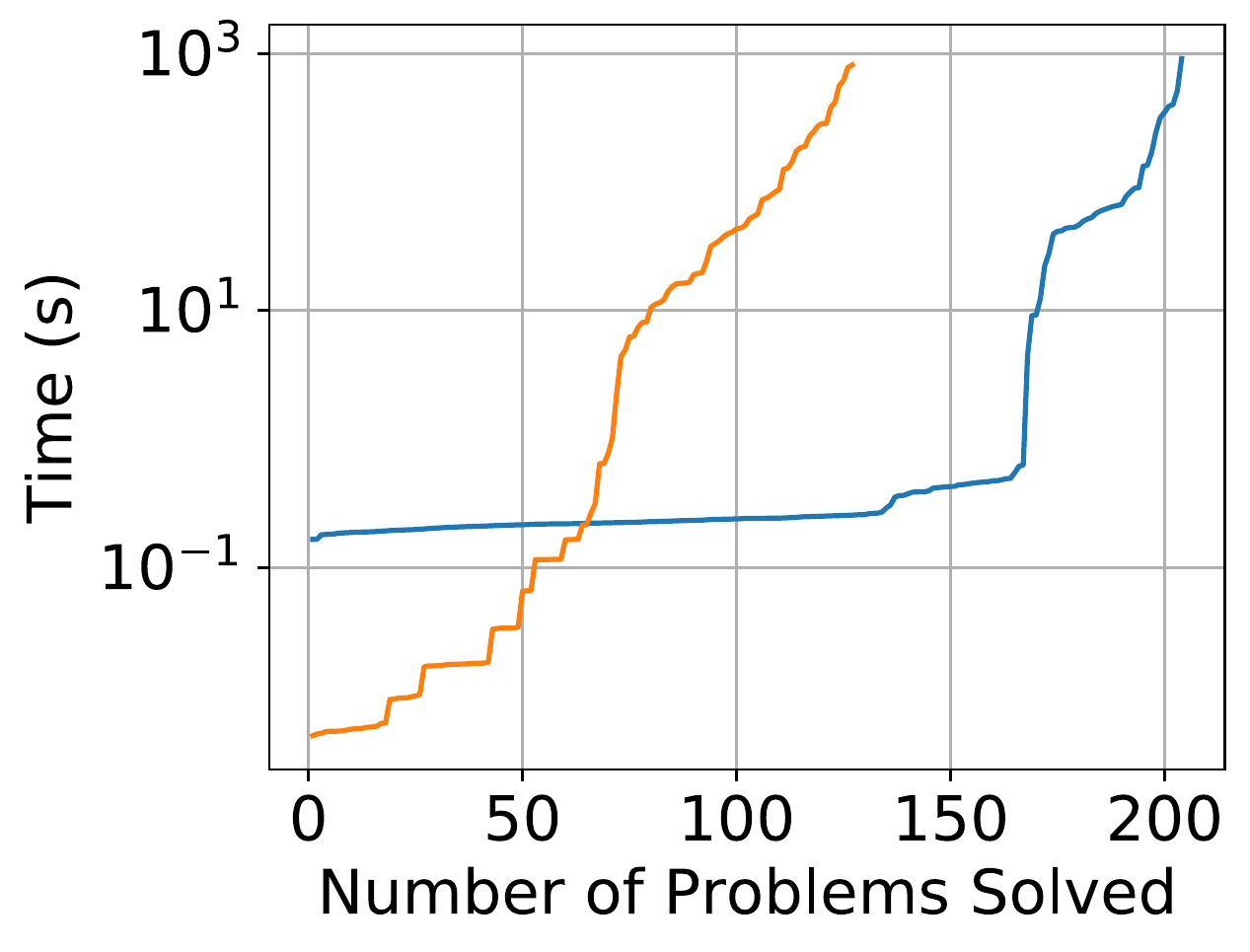}}
    \subfloat[]{\includegraphics[width=.5\linewidth]{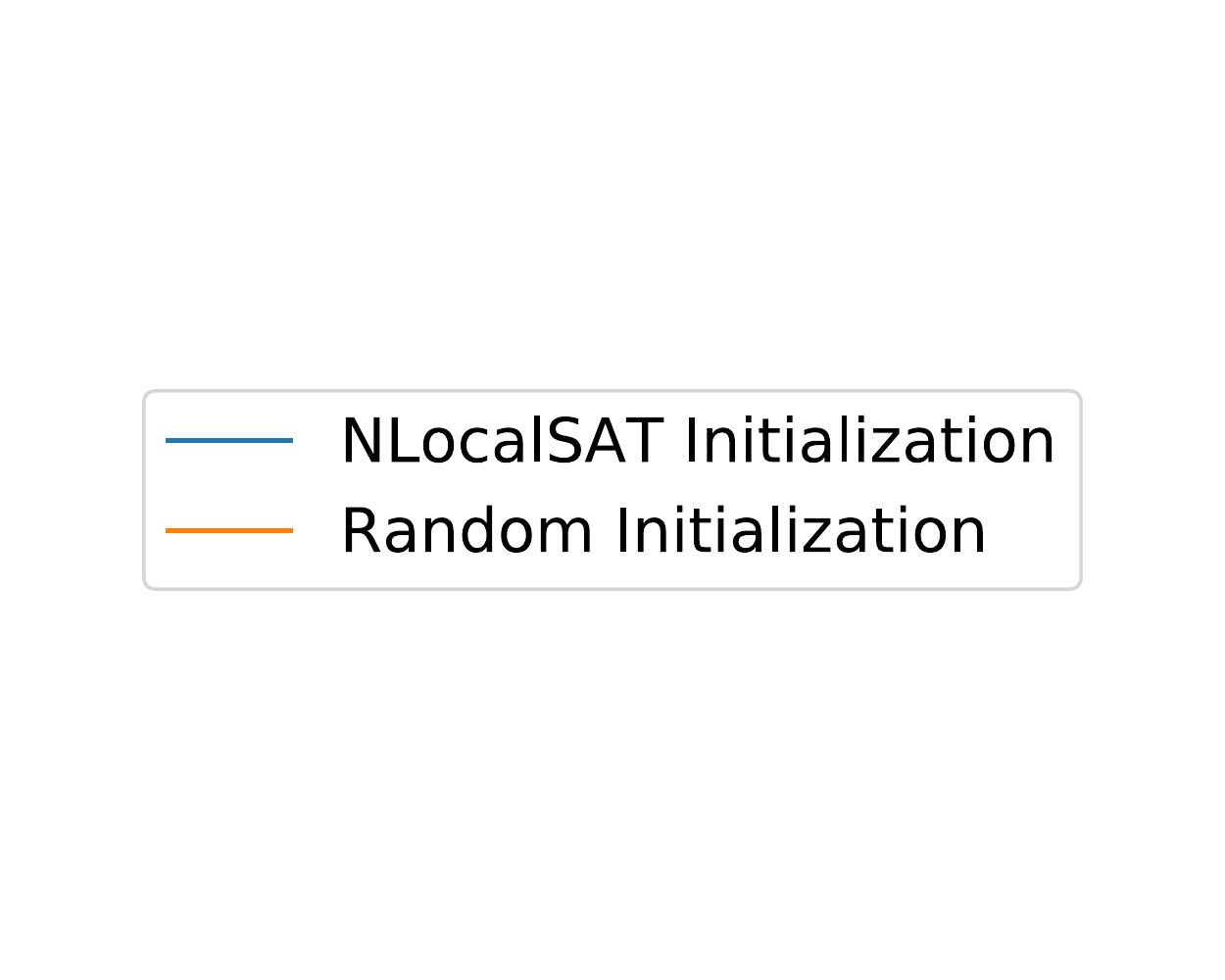}}
    \caption{Cactus plots comparing \techname initialization with random initialization over CCAnr, Sparrow, and CPSparrow.}\label{img:cactus}
\end{figure}

Figure \ref{img:cactus} shows the relationship between the number of instances solved and time consumption comparing solvers with \techname and without \techname. In this figure, we can see that some simple instances which are solved within 1 second with the original solver need more solving time with \techname than the original solver. This is because the neural network computation takes a certain amount of time before the solver starts and this time is especially noticeable for simple instances. But on hard instances, our modifications can improve the solver significantly.

Table \ref{tab:speedup-or-slowdown} compares the average running time with the timeout penalty (PAR-2) between different solvers. The PAR-2 running time of instances that are not solved by the solver in the time limit is twice of the time limit (i.e., 2000 seconds in our experiments). The PAR-2 score was used in previous SAT Competitions. Values in this table show that solvers with \techname are slightly slower on easy instances but much faster on hard instances, particularly, on \textit{Predefined}. Solvers with \techname can find a solution faster than those without \techname on most instances, i.e., solvers with \techname are more effective than the original ones.

\begin{table}
    \centering
    \resizebox{\columnwidth}{!}{
    \begin{tabular}{lcccc}
        \toprule
        Solver & Predefined($165$) & Uniform($90$) & Total($255$)\\
        \midrule
        CCAnr & $747\pm15$ & $1644\pm11$ & $1063\pm7$\\
        CCAnr with \techname & $0.30\pm0.00$ & $1736\pm16$ & $\mathbf{613\pm5}$\\\midrule
        Sparrow & $472\pm7$ & $1531\pm24$ & $846\pm3$\\
        Sparrow with \techname & $0.26\pm0.00$ & $1379\pm9$ & $\mathbf{487\pm3}$\\\midrule
        CPSparrow & $457\pm11$ & $1454\pm38$ & $809\pm13$\\
        CPSparrow with \techname & $0.35\pm0.06$ & $1385\pm9$ & $\mathbf{489\pm3}$\\\midrule
        YalSAT & $1095\pm3$ & $962\pm33$ & $1048\pm9$\\
        YalSAT with 	\techname & $0.26\pm0.00$ & $1226\pm23$ & $\mathbf{433\pm8}$\\\midrule
        probSAT & $1086\pm5$ & $928\pm3$ & $1030\pm3$\\
        probSAT with 	\techname & $0.26\pm0.00$ & $1155\pm26$ & $\mathbf{408\pm9}$\\\midrule
        Sparrow2Riss & $107.7$ & $1560$ & $620$\\
        gluHack & $15.0$ & $2000$ & $715$\\
        MapleLCMDistBT & $8.8$ & $2000$ & $711$\\
        \bottomrule     
    \end{tabular}
    }
    \caption{Average running time with timeout penalty (PAR-2).}\label{tab:speedup-or-slowdown}
\end{table}

\subsection{Discussion}

To find out why our method can boost SLS solvers, we analyzed the experimental result on CCAnr. The geometric mean of ratios of steps that the solver took to find a solution with \techname to the steps without \techname is 0.005. Namely, the solver with \techname takes much fewer steps to find a solution overall. Among all solved instances, the average proportion of correctly predicted variables (i.e., those where the predicted value by the neural network and the final value by the solver are the same) is 0.88, and, with a chi-square analysis, this rate and the speedup of our approach shows a correlation with the p-value $3 \times 10 ^ {-5}$. This experimentally verified our intuition that the closer the initial values are to the solution of the instance, the easier the solver can find a solution. Our neural network can give better initial values, which can boost SLS solvers.

\section{Related Work}

Recently, several studies have investigated how to make use of neural networks in solving NP-complete constraint problems. There are two categories of methods to solve NP-complete problems using neural networks. The first category of methods is end-to-end approaches using end-to-end neural networks to solve SAT instances, i.e., given the instance as an input, the neural network outputs the solution directly. In these methods, the neural network can learn to solve the instance itself~\cite{DBLP:conf/iclr/AmizadehMW19,DBLP:conf/cpaior/Galassi0MM18,DBLP:conf/iclr/SelsamLBLMD19,DBLP:conf/cp/0003KK18,DBLP:conf/aaai/PratesALLV19}. However, due to the accuracy and structural limitations of neural networks, the end-to-end methods can only solve small instances. The other category of methods is heuristic methods that treat neural networks as heuristics~\cite{DBLP:conf/nips/BalunovicBV18,DBLP:conf/nips/LiCK18,DBLP:conf/sat/SelsamB19}. Among these methods, traditional solvers work with neural networks together. Neural networks generate some predictions, and the solvers use these predictions as heuristics to solve the instance.  Constraints in the instances can be maintained in the solver, so these methods can solve large-scale instances. Our proposed method (i.e., \techname) belongs to the second category, heuristic methods, so \techname can be used for larger instances than those end-to-end methods. Balunovic et al.~\cite{DBLP:conf/nips/BalunovicBV18} proposed a method to learn a strategy for Z3, which greatly improves the efficiency of Z3. Li et al.~\cite{DBLP:conf/nips/LiCK18} proposed a model on solving maximal independent set problems with a tree search algorithm. NeuroCore~\cite{DBLP:conf/sat/SelsamB19} is a method to improve CDCL solvers with predictions of unsat-cores. None of these methods is used to boost stochastic local search solvers with solution predictions and none of these is an off-line method to boost SAT solvers. The training data used in \techname are solutions of instances or solution space distribution of instances, which is also different from previous works, where NeuroSAT uses the satisfiability of instances and NeuroCore uses unsat-core predictions.

\section{Conclusion and Future Work}

This paper explores a novel perspective of combining SLS with a solution prediction model. We propose \techname to boost SLS solvers. Experimental results show that \techname significantly increases the number of instances solved and decreases the solving time for hard instances. In particular, Sparrow and CPSparrow with our proposed \techname perform better than state-of-the-art CDCL, SLS, and hybrid solvers on the random track of SAT Competition 2018. 

\techname can boost SLS SAT solvers effectively. With this learning-based method, we may build a domain-specific SAT solver without expertise in the domain by training \techname with SAT instances from the domain. It is also interesting to further explore building domain-specific solvers with \techname.

\section*{Acknowledgements}
This work is sponsored by the National Key Research and Development Program of China under Grant No.~2017YFB1001803, and National Natural Science Foundation of China under Grant
Nos.~61672045, 61922003 and 61832009. Shaowei Cai is supported  by Beijing Academy of Artificial Intelligence (BAAI), and Youth Innovation Promotion Association, Chinese Academy of Sciences [No.~2017150].

\appendix

\bibliographystyle{named}
\bibliography{ijcai20}

\end{document}